# Prediction of Missing Semantic Relations in Lexical-Semantic Network using Random Forest Classifier

Kevin Cousot[1], Mehdi Mirzapour[1], Waleed Ragheb[1]

[1] Université de Montpellier, LIRMM, CNRS, Montpellier, France
{kevin.cousot, mehdi.mirzapour, waleed.ragheb}@lirmm.fr

**Résumé.** Cette étude porte sur la prédiction de six relations sémantiques manquantes (telles que is_a et has_part) entre deux nœuds de RezoJDM, un réseau lexico-sémantique pour le français. Le résultat de cette prédiction est un ensemble de paires dans lesquelles les premières entrées sont des relations sémantiques et les secondes sont les probabilités d'existence de la relation. En raison de l'énoncé du problème, nous avons choisi d'utiliser un classifieur de forêt aléatoire pour le résoudre. Nous exploitons RezoJDM pour en extraire les relations sémantiques et construire nos données d'entraînement/test. Nous expliquons en quoi les idées développées peuvent être utilisées après l'utilisation de l'approche node2vec dans la phase de feature extraction. Nous montrons finalement comment cette approche conduit à des résultats prometteurs.

**Mots-clés.** Apprentissage automatique, apprentissage supervisé, prédiction de relations sémantiques, classifieur de forêts aléatoires, réseau lexico-sémantique, traitement du langage naturel.

**Abstract.** This study focuses on the prediction of missing six semantic relations (such as is_a and has_part) between two given nodes in RezoJDM a French lexical-semantic network. The output of this prediction is a set of pairs in which the first entries are semantic relations and the second entries are the probabilities of existence of such relations. Due to the statement of the problem we choose the random forest (RF) predictor classifier approach to tackle this problem. We take for granted the existing semantic relations, for training/test dataset, gathered and validated by crowdsourcing. We describe how all of the mentioned ideas can be followed after using node2vec approach in the feature extraction phase. We show how this approach can lead to acceptable results.

## 1 Introduction

Prediction of semantic relations in natural language texts is an important task in Natural Language Processing and Computational Linguistics/ Semantics. Scaling up semantic analysis on natural language level needs some sort of semantic relations prediction at some point [8]. Obviously, it is not practical to define all the possible relations for all the possible words or nodes in a network. This naturally leads language models to some sort of trade-off for balancing manually definitions of semantic relations in one hand and prediction of missing relations between unconnected words/nodes in the other hand.

This study specifically focuses on missing semantic relations for French language. Current state-of-the-art methodologies [7], [8] suggest different approaches such as proximity of nodes, similarity-based techniques, maximum likelihood and probabilistic models for solving this kind of the problem. In general, our approach relies on supervised machine learning and tries to introduce the best combination of existing graph embeddings technique [3] to solve the problem of semantic relation prediction for French language.

We have used RezoJDM [5] a French lexical-semantic network which have helped us significantly to create our training/test datasets. To best of our knowledge, no study has been focused on RezoJDM for semantic relation predictions on French vocabularies. The motivation behind this paper is grounding a feasibility study in feature extraction and semantic link predictions on the level of RezoJDM the biggest and state-of-the-art French lexical-semantic network using data-oriented approaches.

Famous supervised machine learning approaches match quite well for the mentioned trade-off that can be re-interpreted as the size of training/test data and the unseen data. To build our training/test dataset, two general approaches can be considered: (i) to manually label the existing corpora with semantic relations (ii) to use state-of-the-art French lexical-semantic networks such as RezoJDM (section 2) which are already built up by crowdsourcing and automatic validations. We focus only on semantic relations; precisely speaking, in this phase of research, we have restricted our focus on the six most important relations, namely: r_raff_sem(relation which links a polysemous term to its different senses), r_syn(relation of synonymy), r_isa (relation of hypernymy, i.e., linking a generic term to a specific instance of it), r_hypo (relation of hyponymy), r_lieu(location relation, i.e., the place where an action / an object can take place / is usually found), r_agent(agent relation) and r_agent-1(inverse of r_agent).

The inputs of our model are set of pairs representing two nodes/words with missing semantic relations in JDM and the output of this prediction is a set of pairs in which the first entries are semantic relations and the second entries are the probabilities of existence of such relations for each pair in the input set. Due to the statement of the problem we choose the random forest (RF) predictor classifier approach [1], [2] to tackle this task. We take for granted the existing semantic relations in RezoJDM, for training/test dataset, gathered and validated by crowdsourcing.

The rest of the paper is organized as follow: Section 2 is a short introduction on RezoJDM. Section 3 is about three-fold training/test datasets building methodology and its feature extraction. Section 4 explains our RF modeling, its parameters setting and the accuracy of the model. In the last section, we conclude our paper and discuss possible future works.

It is worth mentioning that the datasets, saved graphed embedding and python implementation of our RF model is accessible here:
https://github.com/mehdi-mirzapour/JDM_Graph_Embedding

## 2      RezoJDM : A French Lexical-Semantic Network

RezoJDM (http://www.jeuxdemots.org) is a lexical-semantic network that is built by means of online games, so-called Games With A Purpose (GWAP), launched from 2007 [5]. It is a large graph-based network, in constant evolution, containing more than 310,000 terms connected by more than 6.5 million relations. JDM is a graph structure composed of nodes (vertices) and links. Each node in JDM is a 3-tuple: <name, type, weight> and the link is a 4-tuple: <start-node, type, end-node, weight>.

The name is a string denoting the term. The node type is an encoding referring to the information of the node. For instance, a node can be a term or a Part of Speech. The link type refers to the relation considered. A node weight is interpreted as a value referring to the frequency of usage of the term. The relation weight, similarly, refers to the strength of the relation. The link types are divided into predetermined list of the categories, namely: lexical, ontological, associative and predicative relations.

Validation of the relations in RezoJDM which are anonymously given by a player is made also anonymously by other players. A relation is considered valid if it is given by at least one pair of players. A game happens between two players. A first player-- let's say player A-- begins a game with prompting an instruction with a term T randomly picked from the database. The player A has then a limited time to answer which answer is applicable to the term T. The number of propositions which he can make is limited to letting the player not just type anything as fast as possible, but to have to choose the proper one. The same term with the same instruction is later proposed to another player B; the process is then identical. To increase the playful aspect, for any common answer in the propositions of both players, they receive a given number of points. At the end of a game, propositions made by the two players are shown, as well as the intersection between these terms and the number of points they win.

## 3      Building Training/Test Datasets from RezoJDM Graph

In order to build training/test datasets from RezoJDM, we have taken following three stages:

### 3.1     Subgraphs Selection from RezoJDM

Since JDM is a very big network, we have selected only a part of the existing subgraphs of the network for our training/test dataset in this study. The nodes are restricted just to 2 types of "n_term" and "n_form". The edges are restricted to types of "r_raff_sem", "r_syn|", "r_isa", "r_hypo", "r_lieu", "r_agent-1", "r_agent". Experimental trials suggest the threshold of 60 for the weight attributes of all the mentioned relations.

### 3.2     Feature Learning from RezoJDM

For feature learning from RezoJDM network we have used node2vec approach [3] which is a mapping of nodes to a low-dimensional space of features that maximizes the likelihood of preserving network neighborhoods of nodes. Node2vec defines a flexible notion of a node's network neighborhood and design a biased random walk procedure, which efficiently explores diverse neighborhoods.

Experiments demonstrate that node2vec outperforms state-of-the-art methods by up to 26.7% on multi-label classification and up to 12.6% on link prediction [3]. In this paper, we have avoided to go through technicalities in the node2vec approach. The parameter settings used for node2vec are:

Number of dimensions=128
Length of walk per source= 80
Number of walks per source=1
Return hyperparameter=1
In-out hyperparameter=1
Vector dimension size=20

After applying the node2vec algorithm, we ended up creating vectors with size parameter of 20 for each source and end node in the selected subgraphs. Consequently, 40 columns were built that can properly serve as the input for the further supervised learning algorithm. A cursory investigation on the outcome of our parameter settings shows interesting results: similarity between femme(=woman), homme(=man) is 0.9395609, also the most similar vector to 'femme'(=woman)-'roi'(=king)-'homme'(=man) is 'princesse' (0.99). We will see more in section 4 how our parameter setting can lead to interesting results.

### 3.3  Creating the Training/Test Datasets

We can build a training/test dataset with source and destination words and the relevant ids plus 40 features learned from node2vec approach (section 3.2). There are six columns related to the labels which include binary semantic relations (section 3.1). The labels show the existence/nonexistence of the relations in the JDM network. We have selected only the edges with weights greater than 80 in final stage. We have end up with around 37,000 rows of data for this phase of our study. It is worth mentioning that the cleaned dataset can be accessed here:
(https://github.com/mehdi-mirzapour/JDM_Graph_Embedding/blob/master/cleaned_JDM_dataset.xlsx)

### 4  Prediction of Missing Semantic Relations by means of Random Forest Classifier

Random Forest (RF) algorithm [1], [2] has been considered as one of the most precise prediction methods for classification and regression. In this section we apply this approach for predicting the semantic relations. We finish with our parameter settings and its performance analysis.

### 4.1 Parameter Settings for Random Forest Classifier

Random forest (RF) classifier algorithm [1], [2] is an ensemble-learning algorithm that combines the predictions of multiple models to create a more accurate final prediction. There are a number of parameters that needs to be set for our model that we can tweak them. Some of them are related to the individual trees in the model, and change how they are constructed such as *min_samples_leaf, min_samples_split and max_depth, max_leaf_nodes*. We have set all of these parameters to their default values. There are also parameters specific to the random forest that alter its overall construction such as *n_estimators* and bootstrap (or bagging) which is turned on. For our RF Classifier, the *n_estimators* parameter value is optimized using cross-validation to find the values that could best predict the semantic relation status. For each stage, *n_estimators* from 10 to 170 with intervals of length 20 were tested. The following figure shows that the accuracy of model will not increase a lot after setting *n_estimators* to 170. In real practice, we have set this parameter to 70 for all six type of relations to reduce computing.

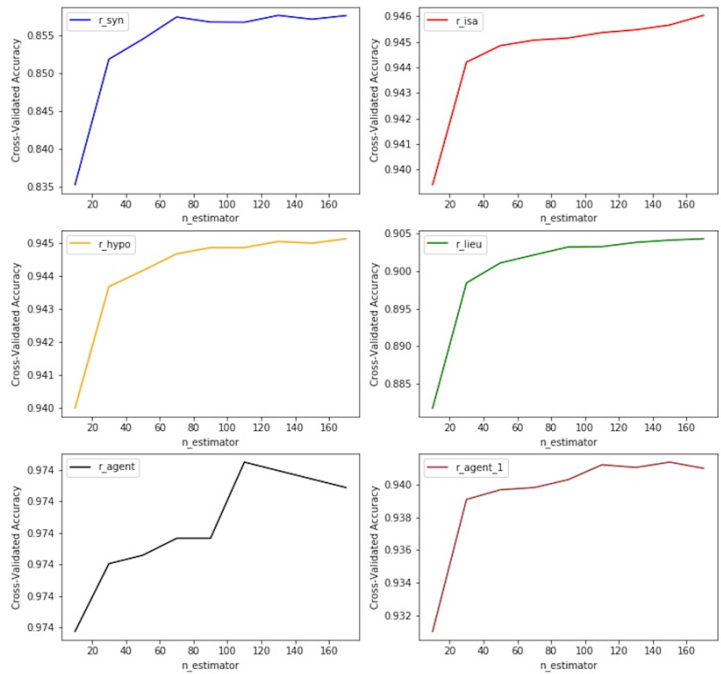

**Fig. 1.** RF n_estimator parameter setting. The analysis shows that setting the parameter to 100 can properly make the trade-off between performance and accuracy.

## 4.2 Performance Analysis

The RF algorithm have several advantages comparing to other frameworks [1]. To count some of them: (i) it runs efficiently on large datasets, (ii) it is less sensitive to noise or over-fitting, (iii) it has fewer parameters, and finally (iv) it has better performance. In this subsection we explain some of the metrics related to our RF performance. As it is illustrated in the table (1), recall, precision and F1 scores have good level of qualities for r_syn, r_is, r_lieu and r_agent_1. It is worth mentioning that our reference for following results is based on JDM test-set itself. This might not be considered as a highly reliable approach and can be interpreted as some sort of internal bias. Nevertheless, this can be counted into account in other studies that deals with reliability problem and it is not explored in this study.

**Table 1.** Recall, Precision, Accuracy and F1 scores after applying the tuned Random Forest for six semantic relations.

|  | r_syn | r_isa | r_hypo | r_lieu | r_agent_1 | r_agent |
|---|---|---|---|---|---|---|
| **Recall** | 0.79 | 0.54 | 0.37 | 0.84 | 0.61 | 0.20 |
| **Precision** | 0.84 | 0.90 | 0.82 | 0.87 | 0.84 | 0.77 |
| **F1 Score** | 0.81 | 0.66 | 0.51 | 0.85 | 0.70 | 0.31 |

Two remaining relations, namely, r_hypo and r_agent do not show proper recall indicators. This has affected the F1 score, since F1 relies on recall as one of its input factors. Low recall and high precision in our modeling have happened due to the unbalanced classes in our datasets. Tweaking *class_weigh* parameter does not improve much our model. Making balanced datasets needs more computing power and better strategies in our dataset creating phase.

## 5 Conclusion and Possible Future Works

In this paper, we reported the feasibility study of prediction of missing semantic relations between two given nodes in RezoJDM French lexical-semantic network. We have also provided a methodology for our feature extraction that is very useful for different natural language processing tasks—regardless of its application for our specific task. We implemented all of the ideas in this paper and we have gained rather satisfactory results: recall 55% and precision 84%, F1 score 64% (percentages are on average for all six types of semantic relations).

We reported an important issue in section 4.2 regarding lacking the balanced datasets. Further enhancement for overcoming this drawback has left for our future studies and needs more consideration on new strategies and balancing the computing resources.

We are planning to extend our study to include the full RezoJDM network with more relations. Moreover, we can apply deeper graph embedding methods to produce unsupervised node features from big graphs and also make use of deeper models rather than the classical machine learning for classification step. Reformulating the problem from binary classification to multilabel and multiclass classification can be done as another extension to our current work.

The introduced workflow in this paper, is intentionally designed for some future projects in computational semantics field. The partial success in this study can pave the way for following future applications. To count some of them, we can mention the following NLP tasks:

(i) In some sentence completion task algorithm [9], we need to eliminate different candidates for completion of a given sentence. One way of doing that is eliminating the candidates that have very low probabilities between the missing words and the existing words in the sentence.

(ii) Finding linguistic coercions in computational linguistics field [6], [10] is another important task for deep semantic analysis and semantic preference modeling.

(iii) Measuring linguistic complexity on syntactic level [11] needs to be enriched with some extra semantic relations on the word level. This task can be definitely enhanced by using semantic relations prediction as we have described in this research.

**Acknowledgments.** We would like to appreciate Mathieu Lafourcade, Michael Carl and Christian Retoré for their kind supports and discussions.